\documentclass{caseib2026}

\usepackage{booktabs}
\usepackage{multirow}
\usepackage{tabularx}
\usepackage{arydshln}
\usepackage{xcolor}
\usepackage{colortbl}
\usepackage{fontawesome}
\usepackage{pifont}
\newcolumntype{C}{>{\centering\arraybackslash}X}
\definecolor{darkgreen}{RGB}{0, 150, 0}
\definecolor{darkred}{RGB}{200, 0, 0}
\newcommand{\cmark}{\ding{51}}
\newcommand{\xmark}{\ding{55}}
\usepackage{graphicx}
\usepackage{subcaption}
\usepackage{marvosym}

\definecolor{lime}{HTML}{A6CE39}

\title{Do Center Biases Propagate? Robustness of Pathology Foundation Models in Whole-Slide Image Classification}
\author{Ilán Carretero$^{1}$, Pablo Meseguer$^{1}$, Rocío del Amor$^{1,2}$, and Valery Naranjo$^{1,2}$}
\affiliations{%
$^1$ CVBLab, HumanTech, Universitat Politècnica de València (UPV), Valencia, Spain\\
$^2$ Artikode Intelligence S.L., Valencia, Spain\\
\Letter\ \{ilcarjuc, pabmees, madeam2, vnaranjo\}@upv.es
}

\shorttitle{Robustness of Pathology Foundation Models}
\shortauthors{Carretero, I. et al.}

\howtocite{%
How to cite: Carretero, I.; Meseguer, P.; Del Amor, R.; and Naranjo, V. (2026). Do Center Biases Propagate? Robustness of Pathology Foundation Models in Whole-Slide Image Classification. En libro de actas: \textit{XLIV Congreso Anual de la Sociedad Española de la Ingeniería Biomédica 2026}. Valencia, 11--13 November 2026. \href{https://doi.org/10.4995/CASEIB2026.2026.xxxxx}{https://doi.org/10.4995/CASEIB2026.2026.xxxxx}
}

\begin{document}

\maketitle

\begin{abstract}
Pathology foundation models (PFMs) have transformed computational pathology through powerful representation learning from histopathological images. PFMs provide rich, discriminative representations for whole slide image (WSI) analysis, enabling tasks such as slide-level classification under multiple instance learning (MIL). However, these representations may also encode non-biological signals associated with acquisition centers, potentially introducing spurious shortcuts into downstream predictions. In this work, we evaluate center-associated robustness in WSI classification using a controlled training setting with increasing class-center correlations quantified by Cramér's $V$. We benchmark six PFMs across four datasets and two MIL aggregators, while evaluating ComBat as a robustification strategy. We further introduce the Area Under the Cramér's $V$ Curve (AUCC) to jointly capture absolute classification performance and its degradation as spurious correlation increases. Results show that center-related information encoded by PFMs propagates to WSI-level predictions, with robustness depending on both the PFM representation and MIL aggregation strategy. Additionally, ComBat harmonization does not provide consistent robustness gains across datasets.
\end{abstract}

\keywords{Pathology Foundation Models, Center Robustness, Whole Slide Images, Spurious Correlations.}

\section{Introduction}
The emergence of pathology foundation models (PFMs) have notably enhanced AI-based workflows in computational pathology (CPath) \cite{ochi2025pathology}. Due to high-capacity model architectures and large-scale, self-supervised, in-domain pretraining, PFMs extract powerful data representations from histopathological images. Their enhanced representational learning has enhanced CPath algorithms for tasks such as whole slide image (WSI) classification. Given the gigapixel scale of WSIs, multiple instance learning (MIL) has become the most prominent approach for WSI classification, requiring only WSI-level annotations \cite{campanella2019clinical}. 

Beyond disease-related biological patterns, PFMs have been shown to encode signals associated with center-specific procedures \cite{komen2026towards}. Across pathology departments, differences in tissue processing, staining protocols, and digitization scanners can introduce variations in color, intensity, and other image characteristics. These technical signatures may act as spurious shortcuts, causing downstream models to rely on center-related information rather than biological signal and limiting generalization across centers. Although large-scale in-domain pretraining may improve robustness to such shifts, its impact on clinically relevant downstream predictions remains an important consideration.

In this work, we evaluate center-associated robustness in WSI-level prediction using a weakly supervised MIL framework. We construct controlled training settings with increasing class--center correlations quantified by Cramér's $V$ and assess how these spurious associations propagate to downstream predictions across PFMs and MIL aggregators. We further introduce the Area Under the Cramér's $V$ Curve (AUCC), which jointly captures absolute classification performance and performance degradation as center--label correlation increases. Finally, we evaluate ComBat harmonization \cite{johnson2007adjusting} as a robustification strategy for mitigating center-associated variation in PFM representations.

\section{Methodology}
\begin{figure*}[htbp]
    \centering
    \includegraphics[width=0.9\textwidth]{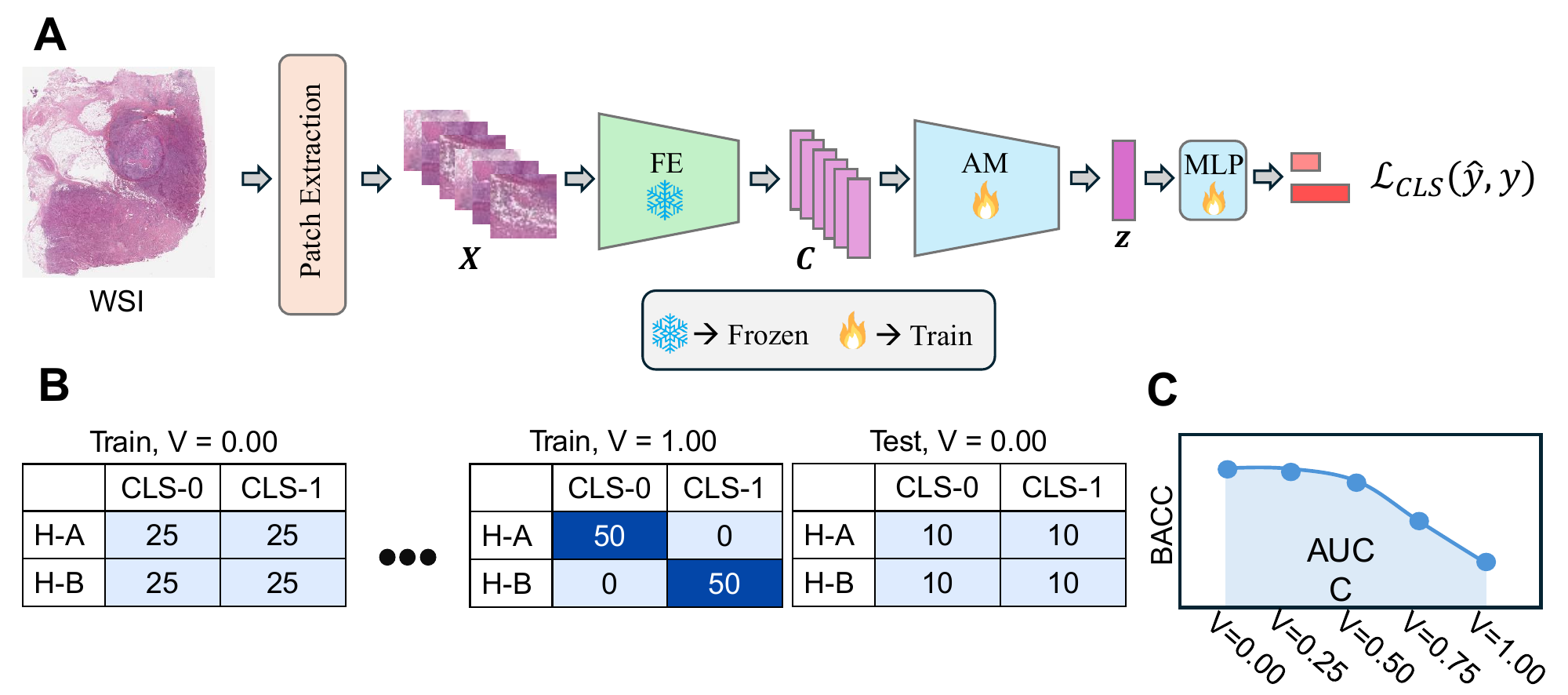}
    \caption{Framework overview: (A) MIL-based WSI classification, (B) center-label spurious correlations, and (C) Robustness quantification.}
    \label{fig:methods}
\end{figure*}

This section presents our framework for evaluating center-associated robustness of PFMs under spurious correlations. Figure~\ref{fig:methods} summarizes the overall approach and its main components are detailed below.

\subsection{Multiple Instance Learning (MIL)}

In MIL, each WSI is represented as a bag $X=\{x_n\}_{n=1}^{N}$ of $N$ image patches and assigned a slide-level label $y\in{Y}$, where ${Y}$ denotes the set of target classes. A PFM encoder maps each patch into a feature representation in $c$, which is subsequently aggregated by a trainable MIL module into a slide-level embedding $z$. Finally, an MLP maps $z$ to the prediction $\hat{y}$, and the downstream model is optimized using the cross-entropy loss $\mathcal{L}_{CE}$.

\subsection{Center-Label Spurious Correlations}
To assess reliance on center-related shortcuts, we construct training sets with progressively stronger associations between the class label $y \in Y$ and acquisition center $h \in H$, while keeping the number of WSIs fixed. The association is quantified through Cramér's $V$,
\vspace{1em}
{\small
\[
V = \sqrt{\frac{\chi^2}{M\min(|Y|-1,\,|H|-1)}},
\quad
\chi^2 = \sum_{y\in Y}\sum_{h\in H}
\frac{(O_{yh}-E_{yh})^2}{E_{yh}},
\]
}

where $M$ is the number of training WSIs, and $O_{yh}$ and $E_{yh}$ are the observed and expected counts for each class-center pair. As shown in Figure \ref{fig:methods} B, Cramér's $V\in[0,1]$ ranges from independence ($V=0$) to maximal association ($V=1$). The test set is always kept uncorrelated ($V=0$), such that performance degradation reflects reliance on the spurious center-label association.

\subsection{Robustness Quantification with AUCC}

To summarize downstream robustness across the different center-label correlations, we propose the Area Under the Cramér's $V$ Curve (AUCC) score. Given the ordered configurations $\{V_t\}_{t=0}^{T}$, we compute the area under the balanced-accuracy curve using trapezoidal integration,
\[
\mathrm{AUCC}
=
\sum_{t=1}^{T}
\frac{\mathrm{BACC}(V_t)+\mathrm{BACC}(V_{t-1})}{2}
\left(V_t-V_{t-1}\right),
\]
where $\mathrm{BACC}(V_t)$ denotes the balanced accuracy obtained when training under correlation level $V_t$, and $T+1$ is the number of evaluated configurations. Unlike measuring performance degradation at a single correlation level, AUCC jointly captures absolute predictive performance and its stability as the spurious association increases. Thus, higher AUCC indicates a stronger performance-robustness trade-off across the complete range of center--label correlations.

\section{Experimental design}
\subsection{Datasets and Preprocessing}
We evaluate four binary WSI classification tasks spanning skin (AI4SKIN \cite{del2025fusocelular}), breast (CAMELYON16 \cite{litjens20181399}, TCGA-BRCA) and lung (TCGA-NSCLC) cancer. For each dataset, we retain WSIs from two acquisition centers, enabling controlled manipulation of center-label correlations. Table~\ref{tbl:dataset} summarizes the resulting class and center distributions.

\begin{table}[h]
\centering
\renewcommand{\arraystretch}{1.10}
\setlength{\tabcolsep}{4pt}
\scriptsize
\begin{tabular}{@{} l l c c c @{}}
\toprule
\multirow{2}{*}{\textbf{Dataset}} &
\multirow{2}{*}{\textbf{Class}} &
\multicolumn{2}{c}{\textbf{Center}} &
\multirow{2}{*}{\textbf{Total}} \\
\cmidrule(lr){3-4}
& & \textbf{H-A} & \textbf{H-B} & \\
\midrule

\multirow{2}{*}{AI4SKIN \cite{del2025fusocelular}}
& Benign    & 133 & 166 & 299 \\
& Malignant & 136 & 191 & 327 \\
\midrule

\multirow{2}{*}{CAMELYON16 \cite{litjens20181399}}
& Negative & 150 & 89 & 239 \\
& Positive & 95  & 65 & 160 \\
\midrule

\multirow{2}{*}{TCGA-BRCA}
& IDC & 91 & 74 & 165 \\
& ILC & 18 & 21 & 39 \\
\midrule

\multirow{2}{*}{TCGA-NSCLC}
& LUAD & 84 & 91 & 175 \\
& LUSC & 50 & 41 & 91 \\
\bottomrule

\end{tabular}

\caption{WSI distribution by class and acquisition center. IDC: invasive ductal carcinoma; ILC: invasive lobular carcinoma; LUAD: lung adenocarcinoma; LUSC: lung squamous cell carcinoma.}
\label{tbl:dataset}
\end{table}
\vspace{-1em}
WSIs undergo tissue detection, downsampling to $10\times$ magnification (1 MPP), and tiling into $512\times512$ patches. We deliberately avoid stain normalization to preserve center-associated appearance and apply the input transformations prescribed by each PFM.

\begin{table*}[tb]
\centering

\renewcommand{\arraystretch}{1.15}
\setlength{\tabcolsep}{4pt}
\scriptsize

\begin{tabular}{@{} l l c c c c c c @{}}
\toprule
\multirow{2}{*}{\textbf{PFM}} &
\multirow{2}{*}{\textbf{Aggregator}} &
\multirow{2}{*}{\textbf{ComBat \cite{johnson2007adjusting}}} &
\multicolumn{5}{c}{\textbf{AUCC ($\uparrow$)}} \\
\cmidrule(l){4-8}
& & &
\textbf{AI4SKIN \cite{del2025fusocelular}} &
\textbf{CAMELYON16 \cite{litjens20181399}} &
\textbf{TCGA-BRCA} &
\textbf{TCGA-NSCLC} &
\textbf{Average} \\
\midrule

\multirow{4}{*}{CONCH {\tiny\textcolor{gray}{Nat. Med.'24}} \cite{lu2024visual}}
& \multirow{2}{*}{ABMIL \cite{ilse2018attention}}
& \xmark
& 88.42 & 79.06 & 88.77 & 83.76 & 85.00 \\

&
& \cmark
& 86.71{\tiny\color{darkred}{$\downarrow$1.71}}
& 81.13{\tiny\color{darkgreen}{$\uparrow$2.07}}
& 84.75{\tiny\color{darkred}{$\downarrow$4.02}}
& 83.37{\tiny\color{darkred}{$\downarrow$0.39}}
& 83.99{\tiny\color{darkred}{$\downarrow$1.01}} \\

\cdashline{2-8}

& \multirow{2}{*}{TransMIL \cite{shao2021transmil}}
& \xmark
& 87.54 & 66.19 & 90.52 & 81.37 & 81.40 \\

&
& \cmark
& 86.67{\tiny\color{darkred}{$\downarrow$0.87}}
& 74.03{\tiny\color{darkgreen}{$\uparrow$7.84}}
& 86.10{\tiny\color{darkred}{$\downarrow$4.42}}
& 80.10{\tiny\color{darkred}{$\downarrow$1.27}}
& 81.73{\tiny\color{darkgreen}{$\uparrow$0.33}} \\

\midrule

\multirow{4}{*}{KAIKO {\tiny\textcolor{gray}{arXiv'24}} \cite{aben2024towards}}
& \multirow{2}{*}{ABMIL}
& \xmark
& 82.80 & 61.57 & 61.69 & 63.02 & 67.27 \\

&
& \cmark
& 81.95{\tiny\color{darkred}{$\downarrow$0.85}}
& 62.49{\tiny\color{darkgreen}{$\uparrow$0.92}}
& 60.96{\tiny\color{darkred}{$\downarrow$0.73}}
& 63.05{\tiny\color{darkgreen}{$\uparrow$0.03}}
& 67.11{\tiny\color{darkred}{$\downarrow$0.16}} \\

\cdashline{2-8}

& \multirow{2}{*}{TransMIL}
& \xmark
& 82.21 & 57.49 & 59.21 & 62.26 & 65.29 \\

&
& \cmark
& 82.52{\tiny\color{darkgreen}{$\uparrow$0.31}}
& 59.10{\tiny\color{darkgreen}{$\uparrow$1.61}}
& 60.19{\tiny\color{darkgreen}{$\uparrow$0.98}}
& 62.31{\tiny\color{darkgreen}{$\uparrow$0.05}}
& 66.03{\tiny\color{darkgreen}{$\uparrow$0.74}} \\

\midrule

\multirow{4}{*}{VIRCHOW2 {\tiny\textcolor{gray}{arXiv'24}} \cite{zimmermann2024virchow2}}
& \multirow{2}{*}{ABMIL}
& \xmark
& 88.63 & 65.49 & 66.58 & 68.01 & 72.18 \\

&
& \cmark
& 87.16{\tiny\color{darkred}{$\downarrow$1.47}}
& 63.36{\tiny\color{darkred}{$\downarrow$2.13}}
& 65.21{\tiny\color{darkred}{$\downarrow$1.37}}
& 66.79{\tiny\color{darkred}{$\downarrow$1.22}}
& 70.63{\tiny\color{darkred}{$\downarrow$1.55}} \\

\cdashline{2-8}

& \multirow{2}{*}{TransMIL}
& \xmark
& 87.13 & 60.90 & 66.88 & 67.09 & 70.50 \\

&
& \cmark
& 86.57{\tiny\color{darkred}{$\downarrow$0.56}}
& 62.29{\tiny\color{darkgreen}{$\uparrow$1.39}}
& 66.25{\tiny\color{darkred}{$\downarrow$0.63}}
& 66.71{\tiny\color{darkred}{$\downarrow$0.38}}
& 70.46{\tiny\color{darkred}{$\downarrow$0.04}} \\

\midrule

\multirow{4}{*}{KEEP {\tiny\textcolor{gray}{Cancer Cell'24}} \cite{zhou2026knowledge}}
& \multirow{2}{*}{ABMIL}
& \xmark
& 81.99 & 54.76 & 85.69 & 83.35 & 76.45 \\

&
& \cmark
& 81.08{\tiny\color{darkred}{$\downarrow$0.91}}
& 59.82{\tiny\color{darkgreen}{$\uparrow$5.06}}
& 77.21{\tiny\color{darkred}{$\downarrow$8.48}}
& 79.56{\tiny\color{darkred}{$\downarrow$3.79}}
& 74.42{\tiny\color{darkred}{$\downarrow$2.03}} \\

\cdashline{2-8}

& \multirow{2}{*}{TransMIL}
& \xmark
& 87.58 & 57.89 & 88.50 & 81.36 & 78.83 \\

&
& \cmark
& 87.13{\tiny\color{darkred}{$\downarrow$0.45}}
& 58.94{\tiny\color{darkgreen}{$\uparrow$1.05}}
& 84.19{\tiny\color{darkred}{$\downarrow$4.31}}
& 80.17{\tiny\color{darkred}{$\downarrow$1.19}}
& 77.61{\tiny\color{darkred}{$\downarrow$1.22}} \\

\midrule

\multirow{4}{*}{UNI2 {\tiny\textcolor{gray}{Nat. Med.'25}} \cite{chen2024uni}}
& \multirow{2}{*}{ABMIL}
& \xmark
& 87.55 & 63.58 & 67.23 & 69.05 & 71.85 \\

&
& \cmark
& 84.59{\tiny\color{darkred}{$\downarrow$2.96}}
& 63.39{\tiny\color{darkred}{$\downarrow$0.19}}
& 65.90{\tiny\color{darkred}{$\downarrow$1.33}}
& 68.80{\tiny\color{darkred}{$\downarrow$0.25}}
& 70.67{\tiny\color{darkred}{$\downarrow$1.18}} \\

\cdashline{2-8}

& \multirow{2}{*}{TransMIL}
& \xmark
& 87.49 & 62.11 & 65.27 & 66.39 & 70.31 \\

&
& \cmark
& 86.24{\tiny\color{darkred}{$\downarrow$1.25}}
& 63.30{\tiny\color{darkgreen}{$\uparrow$1.19}}
& 64.06{\tiny\color{darkred}{$\downarrow$1.21}}
& 65.73{\tiny\color{darkred}{$\downarrow$0.66}}
& 69.83{\tiny\color{darkred}{$\downarrow$0.48}} \\

\midrule

\multirow{4}{*}{HOptimus1 {\tiny\textcolor{gray}{AACR'25}} \cite{scalbert2026hoptimus1}}
& \multirow{2}{*}{ABMIL}
& \xmark
& 81.89 & 56.16 & 65.21 & 67.41 & 67.67 \\

&
& \cmark
& 81.14{\tiny\color{darkred}{$\downarrow$0.75}}
& 56.03{\tiny\color{darkred}{$\downarrow$0.13}}
& 63.17{\tiny\color{darkred}{$\downarrow$2.04}}
& 66.44{\tiny\color{darkred}{$\downarrow$0.97}}
& 66.69{\tiny\color{darkred}{$\downarrow$0.98}} \\

\cdashline{2-8}

& \multirow{2}{*}{TransMIL}
& \xmark
& 82.57 & 54.70 & 63.00 & 67.71 & 67.00 \\

&
& \cmark
& 82.92{\tiny\color{darkgreen}{$\uparrow$0.35}}
& 55.07{\tiny\color{darkgreen}{$\uparrow$0.37}}
& 62.94{\tiny\color{darkred}{$\downarrow$0.06}}
& 67.10{\tiny\color{darkred}{$\downarrow$0.61}}
& 67.01{\tiny\color{darkgreen}{$\uparrow$0.01}} \\

\bottomrule
\end{tabular}
\caption{AUCC robustness of patch-level pathology FMs across datasets and MIL aggregators. Arrows denote the absolute percentage-point change after ComBat correction. Model years indicate the initial public release of each PFM, rather than the publication year of the associated reference.}
\label{tab:main_results}
\end{table*}

\subsection{Models and Evaluation Protocol}

We evaluate six frozen PFMs: CONCH \cite{lu2024visual} and KEEP \cite{zhou2026knowledge} as vision-language models, and VIRCHOW-2 \cite{zimmermann2024virchow2}, H-Optimus-1 \cite{scalbert2026hoptimus1}, KAIKO \cite{aben2024towards}, and UNI-2 \cite{chen2024uni} as vision-only models. Slide-level aggregation uses ABMIL \cite{ilse2018attention} and TransMIL \cite{shao2021transmil}, trained for 20 epochs with AdamW and learning rates of $10^{-4}$ and $10^{-5}$, respectively.

We use 5-fold cross-validation and evaluate five levels of Cramér's $V$, with $V\in\{0,0.25,0.5,0.75,1\}$. Each configuration is repeated over 10 random training-set samplings while keeping the test fold fixed and uncorrelated ($V=0$), with results averaged across folds and samplings.

\section{Results}

\subsection{Representation-Level Center Encoding}

Representation-level center encoding is assessed on mean-pooled slide embeddings using FM-SI \cite{meseguer2025benchmarking} and RI \cite{komen2026towards}. FM-SI measures center-wise clustering (lower is better), whereas RI contrasts biological- and center-driven neighborhood structure (higher is better).

\begin{figure}[h]
    \centering
    \includegraphics[width=\columnwidth]{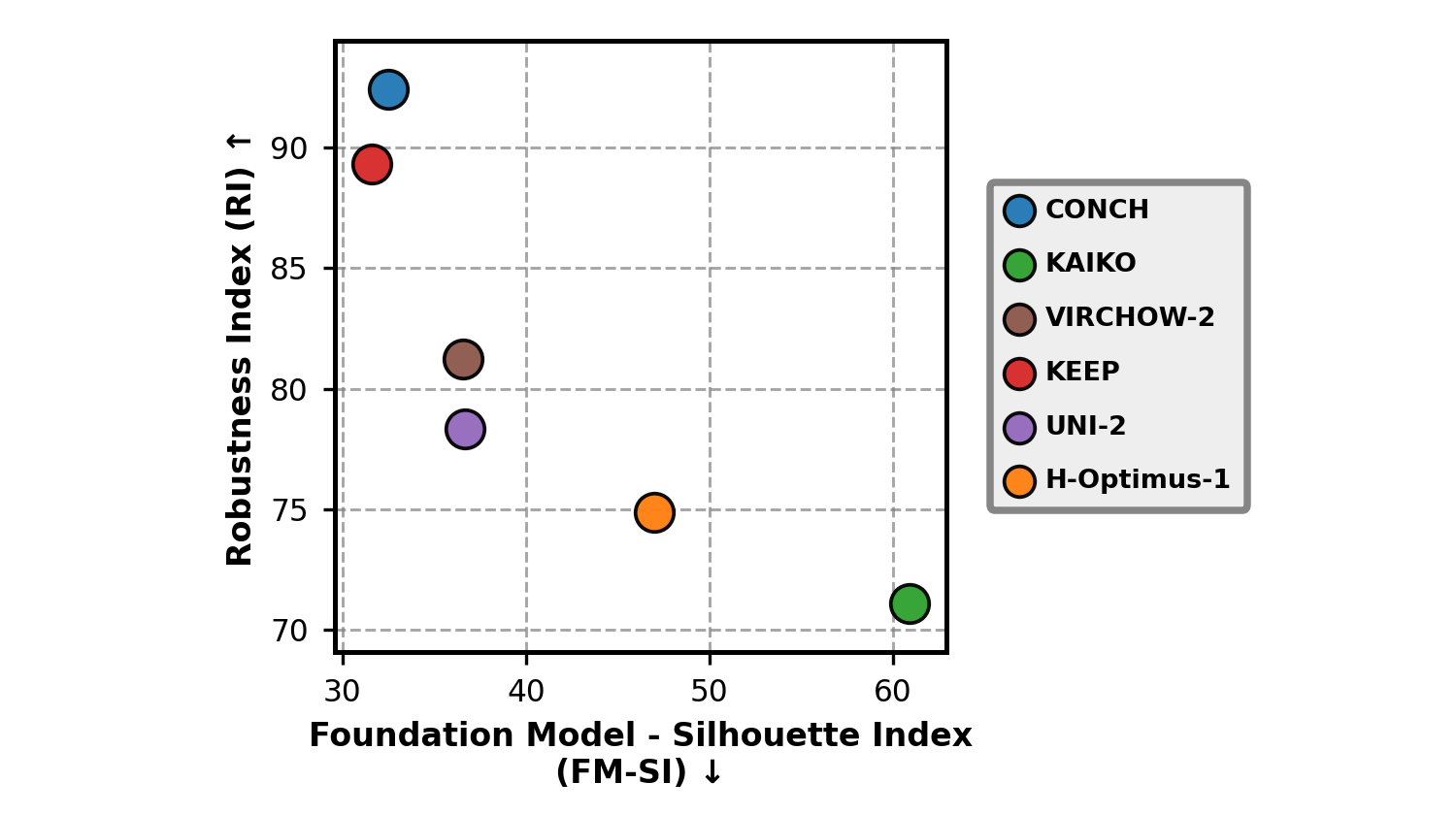}
    \caption{Slide-level robustness of patch-level pathology foundation models across center-related metrics.}
    \label{fig:fmsi_ri}
\end{figure}

Figure~\ref{fig:fmsi_ri} reveals substantial differences in center robustness across PFMs, with lower FM-SI consistently associated with higher RI. CONCH and KEEP occupy the most robust region of the representation space, whereas KAIKO shows the strongest center-related encoding, followed by H-Optimus-1. Notably, both vision-language PFMs outperform all vision-only models on the two complementary metrics, suggesting that, within our benchmark, multimodal pretraining yields representations less dominated by acquisition-center information.

\subsection{Robustness to Spurious Correlations}

Downstream performance is measured by BACC, while AUCC summarizes performance and robustness across the full range of Cramér's $V$.

\begin{figure}[h]
    \centering
    \includegraphics[width=\linewidth]{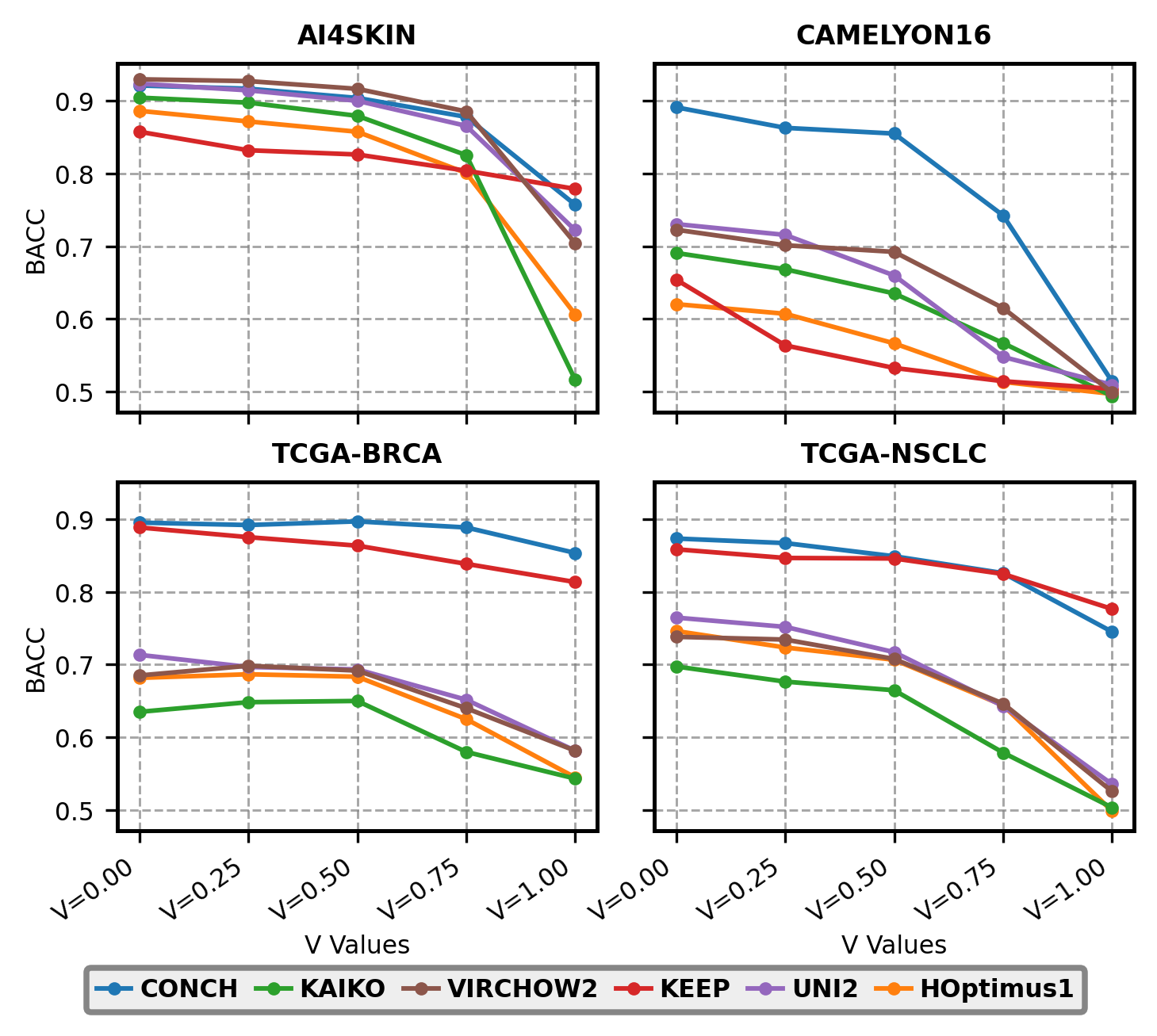}
    \caption{BACC under increasing spurious center--label correlation.}
    \label{fig:cramer_curves}
\end{figure}

Table~\ref{tab:main_results} shows that, without robustification, ABMIL achieves higher average AUCC for five out of six PFMs, outperforming TransMIL by 1.18 percentage points on average. This indicates that the aggregation strategy itself influences downstream robustness, with ABMIL generally preserving performance better under spurious center--label correlations. Figure~\ref{fig:cramer_curves} further reveals a consistent performance degradation as the correlation increases across models and datasets. CONCH and KEEP remain notably robust on TCGA-BRCA at high $V$, whereas most configurations deteriorate substantially, showing that center information encoded in PFM representations can propagate to slide-level predictions through spurious shortcuts.

\subsection{Effect of ComBat Harmonization}

As a widely used harmonization strategy, we apply ComBat \cite{johnson2007adjusting} to mitigate center-associated bias in PFM embeddings. ComBat estimates and removes center-specific additive and scale effects while preserving biological variation.

Despite this correction, ComBat does not provide consistent robustness gains. Across the 48 PFM-aggregator-dataset configurations, AUCC decreases in 34 cases and improves in only 14, yielding an average change of $-0.63$ percentage points. Moreover, its effect is strongly dataset-dependent, with frequent improvements on CAMELYON16 but almost systematic degradation on TCGA-BRCA. These results suggest that generic feature harmonization is insufficient to reliably mitigate center-related shortcuts and may compromise discriminative information in some settings.

\section{Conclusion}

This work provides a controlled evaluation of center-related robustness in pathology foundation models, showing that center information encoded at the representation level can propagate to downstream WSI classification under spurious center--label correlations. Across six PFMs, four datasets, and two MIL aggregators, the proposed AUCC enables a unified assessment of predictive performance and robustness, while ComBat provides limited and inconsistent mitigation of center effects. Our study is nevertheless constrained by the controlled Cramér's $V$ setting, which restricts the available training samples and focuses on binary tasks with two acquisition centers, as well as by the use of two MIL aggregators and patch-level PFMs. Future work will therefore extend this analysis to broader clinical settings and slide-level foundation models, while developing center-robust foundation models and plug-in robustification strategies that can mitigate acquisition-related shortcuts without compromising discriminative information.

\section*{Acknowledgments}

This work has been supported by the Generalitat Valenciana (GVA) through the project CIPROM/2022/20 (PROMETEO) and the Spanish Ministry of Science and Innovation under PID2022-140189OB-C21 (ASSIST).

\printbibliography[title={References}]

@article{lu2024visual,
  title={A visual-language foundation model for computational pathology},
  author={Lu, Ming Y and Chen, Bowen and Williamson, Drew FK and Chen, Richard J and Liang, Ivy and Ding, Tong and Jaume, Guillaume and Odintsov, Igor and Le, Long Phi and Gerber, Georg and others},
  journal={Nature medicine},
  volume={30},
  number={3},
  pages={863--874},
  year={2024},
  publisher={Nature Publishing Group US New York}
}

@article{zhou2026knowledge,
  title={Knowledge-enhanced pretraining for vision-language pathology foundation model on cancer diagnosis},
  author={Zhou, Xiao and Sun, Luoyi and He, Dexuan and Guan, Wenbin and Wang, Ge and Wang, Ruifen and Wang, Lifeng and Yuan, Xiaojun and Sun, Xin and Zhang, Ya and others},
  journal={Cancer Cell},
  volume={44},
  number={4},
  pages={777--791},
  year={2026},
  publisher={Elsevier}
}

@article{zimmermann2024virchow2,
  title={Virchow2: Scaling self-supervised mixed magnification models in pathology},
  author={Zimmermann, Eric and Vorontsov, Eugene and Viret, Julian and Casson, Adam and Zelechowski, Michal and Shaikovski, George and Tenenholtz, Neil and Hall, James and Klimstra, David and Yousfi, Razik and others},
  journal={arXiv preprint arXiv:2408.00738},
  year={2024}
}

@inproceedings{scalbert2026hoptimus1,
  title={H-optimus-1: A foundation model for computational histopathology},
  author={Scalbert, Marin and Saillard, Charlie and Peeters, Thomas and Gonzalez, Liam and Valter, Dasha and Llinares-López, Felipe and Mariet, Zelda E. and Jenatton, Rodolphe},
  booktitle={Proceedings of the American Association for Cancer Research Annual Meeting 2026; Part 2 (Late-Breaking, Clinical Trial, and Invited Abstracts)},
  volume={86},
  number={8\_Suppl},
  pages={LB174},
  year={2026},
  publisher={AACR},
  journal={Cancer Research},
  doi={10.1158/1538-7445.AM2026-LB174}
}

@article{chen2024uni,
  title={Towards a General-Purpose Foundation Model for Computational Pathology},
  author={Chen, Richard J and Ding, Tong and Lu, Ming Y and Williamson, Drew FK and Jaume, Guillaume and Chen, Bowen and Zhang, Andrew and Shao, Daniel and Song, Andrew H and Shaban, Muhammad and others},
  journal={Nature Medicine},
  publisher={Nature Publishing Group},
  year={2024}
}

@article{aben2024towards,
  title={Towards large-scale training of pathology foundation models},
  author={Aben, Nanne and de Jong, Edwin D and Gatopoulos, Ioannis and K{\"a}nzig, Nicolas and Karasikov, Mikhail and Lagr{\'e}, Axel and Moser, Roman and van Doorn, Joost and Tang, Fei and others},
  journal={arXiv preprint arXiv:2404.15217},
  year={2024}
}

@inproceedings{ilse2018attention,
  title={Attention-based deep multiple instance learning},
  author={Ilse, Maximilian and Tomczak, Jakub and Welling, Max},
  booktitle={International conference on machine learning},
  pages={2127--2136},
  year={2018},
  organization={Pmlr}
}

@article{shao2021transmil,
  title={Transmil: Transformer based correlated multiple instance learning for whole slide image classification},
  author={Shao, Zhuchen and Bian, Hao and Chen, Yang and Wang, Yifeng and Zhang, Jian and Ji, Xiangyang and others},
  journal={Advances in neural information processing systems},
  volume={34},
  pages={2136--2147},
  year={2021}
}

@article{del2025fusocelular,
  title={A fusocelular skin dataset with whole slide images for deep learning models},
  author={Del Amor, Roc{\'\i}o and L{\'o}pez-P{\'e}rez, Miguel and Meseguer, Pablo and Morales, Sandra and Terradez, Liria and Aneiros-Fernandez, Jose and Mateos, Javier and Molina, Rafael and Naranjo, Valery},
  journal={Scientific Data},
  volume={12},
  number={1},
  pages={788},
  year={2025},
  publisher={Nature Publishing Group UK London}
}

@article{ochi2025pathology,
  title={Pathology foundation models},
  author={Ochi, Mieko and Komura, Daisuke and Ishikawa, Shumpei},
  journal={JMA journal},
  volume={8},
  number={1},
  pages={121--130},
  year={2025},
  publisher={Japan Medical Association/The Japanese Associaiton of Medical Sciences}
}

@article{campanella2019clinical,
  title={Clinical-grade computational pathology using weakly supervised deep learning on whole slide images},
  author={Campanella, Gabriele and Hanna, Matthew G and Geneslaw, Luke and Miraflor, Allen and Werneck Krauss Silva, Vitor and Busam, Klaus J and Brogi, Edi and Reuter, Victor E and Klimstra, David S and Fuchs, Thomas J},
  journal={Nature medicine},
  volume={25},
  number={8},
  pages={1301--1309},
  year={2019},
  publisher={Nature Publishing Group US New York}
}

@article{komen2026towards,
  title={Towards robust foundation models for digital pathology},
  author={K{\"o}men, Jonah and de Jong, Edwin D and Hense, Julius and Marienwald, Hannah and Dippel, Jonas and Naumann, Philip and Marcus, Eric and Ruff, Lukas and Alber, Maximilian and Teuwen, Jonas and others},
  journal={Nature Communications},
  volume={17},
  number={1},
  pages={5218},
  year={2026},
  publisher={Nature Publishing Group}
}

@article{johnson2007adjusting,
  title={Adjusting batch effects in microarray expression data using empirical Bayes methods},
  author={Johnson, W Evan and Li, Cheng and Rabinovic, Ariel},
  journal={Biostatistics},
  volume={8},
  number={1},
  pages={118--127},
  year={2007},
  publisher={Oxford University Press}
}

@article{litjens20181399,
  title={1399 H\&E-stained sentinel lymph node sections of breast cancer patients: the CAMELYON dataset},
  author={Litjens, Geert and Bandi, Peter and Ehteshami Bejnordi, Babak and Geessink, Oscar and Balkenhol, Maschenka and Bult, Peter and Halilovic, Altuna and Hermsen, Meyke and Van de Loo, Rob and Vogels, Rob and others},
  journal={GigaScience},
  volume={7},
  number={6},
  pages={giy065},
  year={2018},
  publisher={Oxford University Press}
}

@inproceedings{meseguer2025benchmarking,
  title={Benchmarking histopathology foundation models in a multi-center dataset for skin cancer subtyping},
  author={Meseguer, Pablo and del Amor, Roc{\'\i}o and Naranjo, Valery},
  booktitle={Annual Conference on Medical Image Understanding and Analysis},
  pages={16--28},
  year={2025},
  organization={Springer}
}
\end{document}